\documentclass[letterpaper]{article} 
\usepackage{aaai23}  
\usepackage{times}  
\usepackage{helvet}  
\usepackage{courier}  
\usepackage[hyphens]{url}  
\usepackage{graphicx} 
\usepackage{natbib}  
\usepackage{caption} 
\usepackage{algorithm}
\usepackage{algorithmic}

\usepackage{newfloat}
\usepackage{listings}
\DeclareCaptionStyle{ruled}{labelfont=normalfont,labelsep=colon,strut=off} 
\floatstyle{ruled}
\newfloat{listing}{tb}{lst}{}
\floatname{listing}{Listing}
\usepackage{booktabs}
\usepackage{pifont}
\newcommand{\cmark}{\ding{51}}%
\newcommand{\xmark}{\ding{55}}%
\usepackage{siunitx}
\usepackage[version=4]{mhchem}  
\usepackage{amsfonts}

\title{A SWAT-based Reinforcement Learning Framework for Crop Management}
\author {
    Malvern Madondo \textsuperscript{\rm 1},
    Muneeza Azmat \textsuperscript{\rm 2},
    Kelsey DiPietro \textsuperscript{\rm 3},
    Raya Horesh \textsuperscript{\rm 4},
    Michael Jacobs \textsuperscript{\rm 5},
    Arun Bawa \textsuperscript{\rm 6},
    Raghavan Srinivasan \textsuperscript{\rm 6},
    Fearghal O'Donncha \textsuperscript{\rm 4} 
}
\affiliations {
    \textsuperscript{\rm 1} Emory University;
    \textsuperscript{\rm 2} Michigan State university;
    \textsuperscript{\rm 3} Sandia National Laboratories;
    \textsuperscript{\rm 4} IBM Research;\\ 
    \textsuperscript{\rm 5} IBM Corporate Social Responsibility; 
    \textsuperscript{\rm 6} Texas A\&M AgriLife\\
    malvern.madondo@emory.edu, azmatmun@msu.edu, kdipiet@sandia.gov, \{rhoresh@us.,  michael.jacobs@, feardonn@ie.\}ibm.com, \{arun.bawa@ag., r-srinivasan@\}tamu.edu
}

\usepackage{bibentry}

\begin{document}

\maketitle

\begin{abstract}
Crop management involves a series of critical, interdependent decisions or actions in a complex and highly uncertain environment, which exhibit distinct spatial and temporal variations. Managing resource inputs such as fertilizer and irrigation in the face of climate change, dwindling supply, and soaring prices is nothing short of a Herculean task. The ability of machine learning to efficiently interrogate complex, nonlinear, and high-dimensional datasets can revolutionize decision-making in agriculture. In this paper, we introduce a reinforcement learning (RL) environment that leverages the dynamics in the Soil and Water Assessment Tool (SWAT) and enables management practices to be assessed and evaluated on a watershed level. This drastically saves time and resources that would have been otherwise deployed during a full-growing season. We consider crop management as an optimization problem where the objective is to produce higher crop yield while minimizing the use of external farming inputs (specifically, fertilizer and irrigation amounts). The problem is naturally subject to environmental factors such as precipitation, solar radiation, temperature, and soil water content. We demonstrate the utility of our framework by developing and benchmarking various decision-making agents following management strategies informed by standard farming practices and state-of-the-art RL algorithms. 
\end{abstract}

\section{Introduction}
In the face of increasing demand for agricultural products together with growing environmental and economic stressors, farming practices that produce higher yields with minimal resources are becoming a key strategy for climate-smart agriculture \cite{alexandratos2012world, muller2017strategies, ranganathan2018sustainably}. The advent of precision agriculture, an innovative approach to modern farming that combines models and sensor data to improve decision-making, has enabled many farmers to achieve higher crop yields without extending arable land or increasing farming inputs \cite{dutia2014agtech}. Each growing season, farmers make critical operational decisions such as crop selection, planting and harvesting schedules, and application of fertilizers and irrigation. Complexities are exacerbated by a changing climate and the need to achieve global food security while minimizing environmental impacts and carbon emissions, as populations continue to grow. 

The widespread adoption of agricultural technology and a burgeoning big data ecosystem has provided many machine learning researchers and practitioners with opportunities to develop innovative crop and resource management solutions. Increasingly machine learning is being applied to tasks ranging from soil moisture and crop yield prediction to crop disease detection using satellite image data \cite{chlingaryan2018machine, gandhi2022deep, liakos2018machine}. \citet{muller2017strategies} proposed various sustainable food production solutions, including increasing crop production on current arable land, adopting greener farming practices, and reducing food consumption and wastage. We consider the first approach and make contributions toward optimizing crop yields using Reinforcement Learning (RL), a sub-field of machine learning that has proven its utility in decision-making across multiple domains such as healthcare, engineering, and games \cite{sutton2018reinforcement}. 

Agriculture use cases have focused on optimizing crop production subject to various resource constraints such as fertilizer application and water usage \cite{binas2019reinforcement, elavarasan2020crop, overweg2021cropgym, wu2021optimization}. This paper employs RL to autonomously learn the optimal set and distribution of actions to direct crop growth, cognizant of spatial and temporal variations. We leverage the Soil and Water Assessment Tool (SWAT) \cite{arnold1998large, gassman2007soil} numerical simulator to generate representative soil-water-plant synthetic data and to develop an RL environment on which various decision-making agents can be evaluated. Our SWAT-based model incorporates dynamic representations of economic cost, environmental impact, and crop yield. It also produces a reward that characterizes the effects of different agent actions on crop production. By favoring less frequent and minimal amounts of agricultural inputs, it generates a holistic framework to minimize environmental impact and operational costs while still recovering optimal yields. 

\paragraph{Contributions.} Our key contributions are as follows:
\begin{enumerate}
    \item We introduce SWATGym, an OpenAI Gym environment modeled after the widely used Texas A\&M SWAT model \cite{arnold2011soil}. SWATGym simulates crop growth and incorporates geographic and weather data to model the complex soil-water-plant-atmosphere system as described by the SWAT model. 
    \item We provide a simple API to apply custom crop management strategies and/or evaluate them against state-of-the-art RL-based strategies. This presents an opportunity to assess the impact and practicality of a strategy before real-world deployment.
    \item We highlight some promising future research directions in the SWATGym environment, e.g., fine-tuning it to be self-contained and capture the spatial-temporal variability of the seasons.
\end{enumerate}

\section{Background and Related Work}
Nitrogen and water are the primary limiting factors for crop production \cite{rimski2009effect}. Limiting their amounts may hinder crop growth potential while excess amounts are not only costly but can lead to environmental harm downstream (e.g., nutrient loading in rivers and depleted groundwater levels). Further, the optimal amounts are highly sensitive to local variables such as soils and climate. Applying correct amounts of nitrogen fertilizer and irrigation can optimize yields while reducing excess nutrient runoff to local groundwater, river, and lake systems. However, determining the input application rates and schedules that meet environmental and economic goals is not easy. Complexities are amplified by uncertainties such as weather, resource availability, and costs, among other factors. Further, the optimal values are highly dependent on the stage of growth of the crop. As a result, techniques that can inform optimal crop management policy within a complex, dynamic environment are critical to more resilient and sustainable farming methods. Moreover, controlling for the environmental and financial impacts of crop management is central to the tenets of sustainable agriculture in a highly volatile market \cite{binas2019reinforcement, dutia2014agtech}. 

Excessive fertilizer application elevates nitrate levels in the soil. Aided by percolating water, nitrogen may leach down to groundwater or get washed away by surface runoff towards rivers and streams \cite{follett2001nitrogen, sharpley1987environmental}. High nitrate levels are considered harmful to humans and livestock and have been found to affect water quality. Excessive irrigation applications only magnify this problem and wash away fertilizer and other potentially harmful substances into rivers and streams. This leads to environmental degradation, non-optimal crop growth, and economic losses. For sustainability efforts and climate-smart agriculture, models that consider the long-term impact of management practices on water quality are vital.

\subsection{Reinforcement Learning for Crop Management}
Reinforcement learning (RL) involves a decision-making agent interacting with an environment in order to learn a reward-maximizing strategy \cite{sutton2018reinforcement}. RL has only recently been applied to crop management with notable works including \cite{binas2019reinforcement} which explores applying RL to sustainable agriculture as well as \cite{ashcraft2021machine} and \cite{overweg2021cropgym} which optimizes for crop yield subject to irrigation and fertilizer management actions.

\paragraph{Problem formulation.}
We consider an agent interacting with a simulated environment modeled as a finite horizon Markov Decision Process comprising of a tuple $(\mathcal{S}, \mathcal{A}, P, r, \gamma)$ described by the following:
\begin{itemize}
    \item a continuous state space $\mathcal{S}$, consisting of soil, hydrological, and plant data as well as climate inputs.
    \item a continuous action space $\mathcal{A}$, where each action $a$ consists of a tuple of fertilizer and irrigation amounts.
    \item a state transition function $P: \mathcal{S} \times \mathcal{A} \times \mathcal{S} \rightarrow [0, 1]$, which characterizes the probability distribution over states at time $t+1$ given the state and action taken at time $t$.
    \item a measurable reward function $r: \mathcal{S} \times \mathcal{A} \rightarrow \mathbb{R}$, which describes the effect of an agent's choice of actions on crop production.
    \item $\gamma \in [0, 1)$ is the discount factor applied to future returns.
\end{itemize}
The goal is to maximize the cumulative rewards over a finite horizon of length $T$, corresponding to a growing season. At each discrete time step $t \in T$ and given a state $s \in \mathcal{S}$, the agent selects an action $a \in \mathcal{A}$ informed by its policy $\pi: \mathcal{S} \rightarrow \mathcal{A}$. The agent then receives a reward $r$ and a new observation of the environment $s'$. The RL objective is to find the optimal policy $\pi^*$ that maximizes the cumulative reward.

\subsection{Crop Growth models}  
Simulated environments are fast becoming the key to the creation of state-of-the-art algorithms for various learning tasks as evidenced by the success of RL algorithms in games such as Go \cite{silver2016mastering} and robotics. Most existing environments are designed to manage either irrigation operations or fertilizer operations. The most notable environment is the Python Crop Simulation Environment \cite{pcse} which houses various crop models such as WOFOST \cite{van1989wofost} and LINTUL3 \cite{shibu2010lintul3} and has inspired environments such as CropGym \cite{overweg2021cropgym}. One drawback of PCSE is that it requires manual calibration of various model components (soil, crop, weather) as well as the specification of the agro-management activities that will take place.

The CropGym environment simulates winter wheat growth in the Netherlands, with a particular focus on fertilizer management. Other seminal models only focus on irrigation management and include the SIMPLE model \cite{zhao2019simple}, implemented as an OpenAI environment in \cite{ashcraft2021machine} and applied to potato growth simulation in Washington State. \cite{chen2021reinforcement} also focuses on irrigation management and simulates paddy rice growth in China. In contrast to these environments, our proposed environment focuses on both fertilizer and irrigation management and considers processes at the watershed level.

\begin{table*}[hbtp!]
\centering
   \small
   \begin{tabular}{lcr}
   \toprule
   \textbf{Environment} &  \textbf{Fertilizer} & \textbf{Irrigation} \\ 
   \midrule
    \cite{ashcraft2021machine} & \xmark & \cmark\\
    CropGym \cite{overweg2021cropgym} & \cmark & \xmark\\
    \cite{chen2021reinforcement} & \cmark & \xmark\\
    \textbf{SWATGym}  & \cmark & \cmark\\
   \bottomrule
   \end{tabular}
   \caption{Comparison of crop growth reinforcement learning environments.}
\label{table:env_comparison}
\end{table*}

\subsection{SWAT} 
The Soil and Water Assessment Tool simulates physical processes such as crop growth, soil water balance, and nutrient cycling in a watershed \cite{arnold1998large, gassman2007soil}. SWAT primarily considers two production levels: potential production, which represents estimated growth under optimal conditions, and actual production, which is limited by factors such as temperature stress, and nutrient and water availability. SWAT uses a simplified version of the EPIC crop model \cite{williams1989epic} to compute the crop-related variables.

\section{The SWATGym Environment}
The main contribution of our paper is SWATGym, an RL environment based on SWAT that simulates a crop's phenological development, growth, and yield on a daily basis by taking into account the effects of factors such as nutrient cycling, water availability, and temperature. 
\begin{figure}[hbtp!]
\centering
\includegraphics[width=0.9\columnwidth]{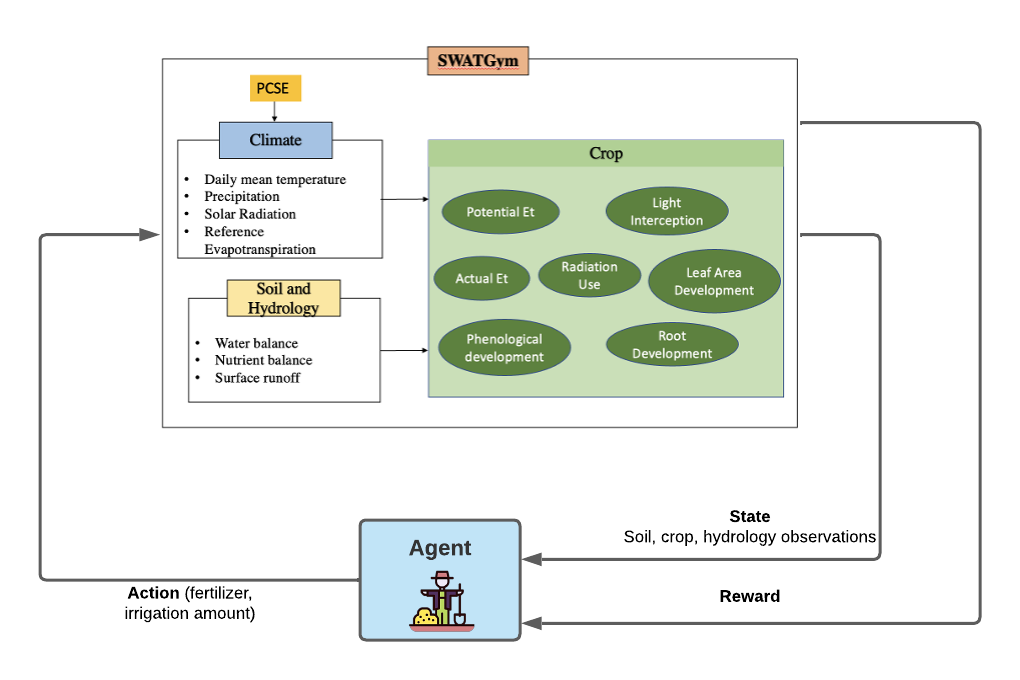}
\caption{The SWATGym environment is the first SWAT-based RL environment. Agents interact with the environment to learn crop management strategies in simulation.}
\label{fig:interface}
\end{figure}

SWATGym is the first Python-based implementation of SWAT, which is written in FORTRAN and is not readily accessible to RL applications. Our environment is built on top of the OpenAI Gym framework, the gold standard framework for developing RL environments \cite{brockman2016openai}. SWATGym primarily considers corn production and incorporates dynamical representations of corn yields and corresponding economic costs of input resources as well as their environmental impact. It has a continuous state space comprising of $14$ state variables describing various processes related to weather, soil, crop, and hydrology dynamics (see Table \ref{table:state_space}). It also has a continuous multidimensional action space. At time $t$, the action is given by $a_t = [F_t, I_t]$, where $F$ and $I$ represent fertilizer and irrigation amounts applied at that time. SWATGym can be run in either stochastic or deterministic modes. The former is enabled by default and is applied primarily through randomness added to climate inputs. The latter mode ensures the same actions lead to the same states.

\begin{table}[!htbp]
\centering
   \small
   \begin{tabular}{lr}
   \toprule
   \textbf{Observation} & \textbf{Unit}\\ 
   \midrule
   Mean air temperature & $\si{\degreeCelsius}$  \\
   Precipitation & mm  \\
   Reference Evapotranspiration & mm\\
   Solar Radiation & $MJ/mm^2$\\
   Mean Vapor Pressure & hPa\\
   Actual Evapotranspiration & mm\\
   Water balance & mm\\
   Daily runoff curve number & - \\
   Leaf area index & -\\
   Nitrogen Uptake & kg/ha\\
   Denitrification & kg/ha\\
   Nitrogen stress factor & - \\
   Temperature stress factor & - \\
   Water stress factor & - \\
   \bottomrule
   \end{tabular}
   \caption{Environment's observations include weather inputs, plant state variables, and soil state variables.}
\label{table:state_space}
\end{table}

\subsection{Reward Function}
 SWATGym produces a reward that characterizes the effect of different choices of actions on crop production. The reward at each time step is computed as 
\begin{equation}
    r_t = yld_t - \alpha F_t - \beta I_t,
\end{equation}
where $yld$ is the estimated crop yield on a particular day and $\alpha$ and $\beta$ are penalty terms associated with the estimated cost of applying fertilizer $F$ and irrigation $I$. In our case, $\alpha = 2.43$ and $\beta = 0.16$.

\section{Crop Growth Simulation}
SWATGym simulates crop growth for a full growing season (about 120 days for corn). Users have the option to specify the location and simulation start date as well as duration. The simulation starts with crop emergence and ends with the harvest. The environment operates on a daily time step. An episode begins on the specified/default simulation date or when the environment is reset and ends after the specified/default duration of the growing season (harvest day). The environment also has the option to save all relevant data about the current growing season, including weather observations, crop states, soil and hydrology balances, actions taken, and yield achieved to date. This feature enables the collection of expert data, which can be used for other tasks such as offline learning. 

\subsection{State-space dynamics}
SWATGym has several major modules such as hydrology, weather, soil temperature, crop growth, and agricultural management \cite{arnold1998large}. It includes state variables describing the time evolution of hydrological, soil, and crop variables. At time (day) $t$, the plant state variables are described by 
\begin{equation}
    z_p(t) = [LAI, BIO, E_a, N_{strs}, W_{strs}, T_{strs}],
\end{equation} and the soil state variables by 
\begin{equation}
    z_s(t) = [SW, RCN, DN, N_{up}].
\end{equation}
$LAI$ is the leaf area index, $BIO$ is the cumulative biomass, $E_a$ is the actual evapotranspiration rate, and $N_{strs}, W_{strs}, \text{ and } T_{strs}$ are growth factors related to stress caused by nitrogen, water, and temperature on the plant. For soil state $z_s$, $SW$ is the available soil water content, $RCN$ is the daily surface runoff curve number, $DN$ is the denitrification rate, and $N_{up}$ is the nitrogen uptake.

We also define a vector of climatic inputs,
\begin{equation}
    \xi(t) = [P, Et, Ta, Rd],
\end{equation}
where $P$ is the precipitation received on that day, $Et$ is the reference evapotranspiration rate, $Ta$ is average daily air temperature,  $Rd$ is daily solar radiation. Therefore, the plant state at time $t+1$ is given explicitly as a nonlinear function of the state and the input climatic variables at time $t$:
\begin{equation}
    z_p(t+1) = f_p\left(z_p(t), z_s(t), \xi(t)\right), \; t \in [0, T], \; z_p(0) = x
\end{equation}
where, for a given initial state $x$, $z_p(t)$ is the vector with the plant state variables at time $t$, $z_s(t)$ is the vector with the soil state variables; and $\xi(t)$ is the vector of the climatic inputs provided by PCSE \cite{pcse} for a specific location and day.

Excluding daily weather data, most of the state variables are propagated by equations provided in the SWAT2009 Theory Documentation \cite{neitsch2011soil}. Below we highlight a few of the crop-related state variables:
\begin{enumerate}
    \item \textbf{Phenology:} Similar to the original SWAT model, we express crop growth/phenological development in terms of heat units, which are driven by daily mean temperature. Growth is accelerated at or above the optimal temperature for the crop, $T_{opt}$, and is slowed or stopped at or below the base temperature $T_{base}$. A crop's phenological development is based on daily heat unit accumulation, given by
    \begin{equation}
        HU_i = \Bar{T}_i - T_{base}, \quad \forall i \in {1, \ldots, T}, \; T>1
    \end{equation}
    where $HU$ is the value of heat units and $\Bar{T}_i$ is the average air temperature in $\si{\degreeCelsius}$ on day $i$. 
    
     The fraction of potential heat units accumulated for a given day $d$ is given by
    \begin{equation}
        fr_{PHU} = \frac{\sum_{i=1}^d HU}{PHU}
    \end{equation}
    where $PHU$ is the total number of heat units required for a plant to reach maturity. This is often calculated from the planting date to the harvest date (on the last day $T$) if not known beforehand.

    Temperature is a key driving factor for crop growth in SWAT. Below we show a plot of observed daily mean temperatures during one simulation run. We also indicate the base and optimal temperatures for corn production.
    \begin{figure}[hbtp!]
        \centering
        \includegraphics[width=0.4\textwidth]{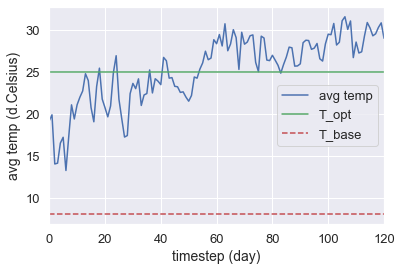}
        \caption{Temperature variation during one growing season, along with the crop-specific base temperature and optimal temperatures for corn production. $T_{base} = 8 \si{\degreeCelsius}$ and $T_{opt} = 25 \si{\degreeCelsius}$}
        \label{fig:temp_variation}
    \end{figure}
    In this example, growth will be limited early on due to suboptimal temperatures. However, other factors such as nitrogen and water balance may encourage growth despite temperature stress.
    
    \item \textbf{Potential Growth:} Other factors related to plant growth that is modeled by SWATGym include leaf area development, light interception, and a plant-specific radiation use efficiency metric (which measures the conversion of intercepted light into biomass). 
    
    \begin{enumerate}
        \item \textit{Leaf Area Development:} The leaf area index (LAI) is the area of green leaf per unit area of land. It is computed as a function of crop canopy height by 
        \begin{equation}\label{eqn:delta_lai}
            \Delta LAI_i = K_f \left(1 - e^{5*(LAI_{i-1} - LAI_{max})}\right)
        \end{equation}
        where $K_f =  LAI_{max}(fr_{LAImax, i} - fr_{LAImax, i-1})$ and $LAI_0 = 0$.
        
        \begin{equation}\label{eqn:lai}
            LAI_i = LAI_{i-1} + \Delta LAI_i
        \end{equation}
        where $h_c$ is the canopy height. For corn, $LAI_{max} = 3 \text{ and } fr_{\text{PHU, sen}}=0.9$ \cite{arnold2011soil}.
        
        \item \textit{Light Interception:} Using Beer's law, the amount of daily solar radiation intercepted by the plant is computed as
        \begin{equation}\label{eqn:light_intercept}
            H_{\text{phosyn}} = 0.5 H_{\text{day}} \left( 1 - exp(-k_{\ell} LAI) \right)
        \end{equation}
        where $H_{\text{phosyn}}$ is the amount of intercepted photosynthetically active solar radiation ($MJ/m^2$), $H_{\text{day}}$ is incident total solar, $k_{\ell}$ is light extinction coefficient and $LAI$ is leaf area index.
        
        \item \textit{Biomass Production:} Radiation Use Efficiency (RUE) is defined for each plant species and is independent of the plant's growth stage. The potential increase in total plant biomass on a given day is given by
        \begin{equation}\label{eqn:potential_bio}
            \Delta bio = RUE \cdot H_{\text{phosyn}}
        \end{equation}
        The total plant biomass on a given day $d$ is subsequently given by 
        \begin{equation}\label{eqn:biomass}
            bio = \sum_{i=1}^d \Delta bio_i, \quad d \leq T
        \end{equation}
    \end{enumerate}
    
    \item \textbf{Crop Yield:} 
    SWATGym computes crop yield as the product of a plant's above-ground biomass (and its roots if they are a harvest-able product) and harvest index, which is defined as the fraction of above-ground plant dry biomass removed as dry economic yield (with values typically between $0$ and $1$). $HI$, the potential harvest index for a given day in the plant's growing season is computed using the following relationship:
    \begin{equation}\label{eqn:hi}
        HI = HI_{\text{opt}} \frac{100 fr_{\text{PHU}}}{100 fr_{\text{PHU}} + exp(11.1 - 10 fr_{\text{PHU}})},
    \end{equation}
    where $HI_{\text{opt}}$ is the potential harvest index at the time of maturity, and $fr_{\text{PHU}}$ is the fraction of potential heat units accumulated for the plant on a given day in the growing season. If needed, the actual harvest index can be derived from Equation \ref{eqn:hi} by taking water deficiency into account.
    
    \begin{figure}[hbtp!]
        \centering
        \includegraphics[width=0.4\textwidth]{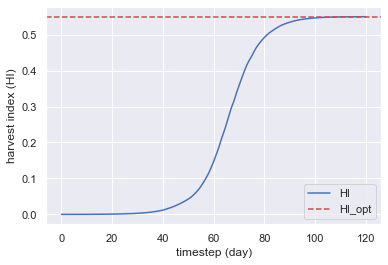}
        \caption{Potential harvest index for corn production during one growing season}
        \label{fig:hi}
    \end{figure}
    
    Overall, the estimated yield is given by
    \begin{equation}
        yld = \begin{cases}
                    bio_{\text{ag}} HI, \quad \text{ for } HI \leq 1     \\
                    bio \left(\frac{HI}{HI+1} \right), \quad \text{ otherwise,}
            \end{cases}
    \end{equation}
    where $yld$ is the crop yield (kg/ha), $bio_{\text{ag}}$ is the above-ground biomass (kg/ha), $HI$ is the harvest index, and $bio$ is the total plant biomass on the day of harvest. $bio_{\text{ag}}$ is computed as follows,
    \begin{equation}
        bio_{\text{ag}} = \left( 1 - fr_{\text{root}} \right) bio,
    \end{equation}
    where $fr_{\text{root}} = 0.4 - 0.2 fr_{\text{PHU}}$ is the fraction of total biomass in the roots on harvest day and $fr_{\text{PHU}}$ is the fraction of potential heat units accumulated for the plant on a given day in the growing season.

        \item \textbf{Growth Constraints:} Plant growth may be affected by insufficient or excess water, nutrients, and extreme temperatures. Stress factors are typically $0$ under normal conditions and approach $1$ as growth conditions become suboptimal. Equations that describe the propagation of these stress factors are provided in \cite{neitsch2011soil}.
    \begin{figure}[hbtp!]
    \centering
    \includegraphics[width=0.4\textwidth]{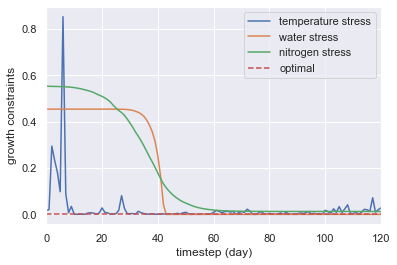}
    \caption{Stress factors during a growing season.}
    \label{fig:stress_factors}
\end{figure}
    \end{enumerate}



\subsection{Source Code}
SWATGym is released as a free and open-source environment under the terms of the Apache License $2.0$. The code is publicly available on GitHub at \url{https://github.com/IBM/SWATgym}. While SWAT accounts for a variety of crop species and environmental processes, we focus on corn production in this paper. We further limit ourselves to key processes such as surface runoff, water balance, and denitrification. For a detailed description of the physiological processes, refer to \cite{neitsch2011soil}.

We provide a simplified implementation of SWAT that makes the following assumptions:
\begin{itemize}
    \item all climatic and agro-management inputs are applied uniformly and daily, over a single growing season (typically 120 days for corn).
    \item all soil layers (except the surface layer, top 10mm) have largely the same characteristics i.e. homogeneous soil profile. 
    \item no/negligible percolation and bypass flow exiting the soil profile at the bottom; no lateral and base flow (which are typically included when computing soil water content).
    \item no growth-reducing factors such as weeds and pests and negligible growth impact of all other nutrients besides nitrogen.
\end{itemize}

\section{Experiments}
In addition to providing the SWATGym environment, we also evaluate a selection of crop management strategies on the environment to demonstrate that RL agents can learn useful crop management strategies. We benchmark the following agents:
\begin{itemize}
    \item \textbf{DDPG.} Deep Deterministic Policy Gradient is a state-of-the-art RL algorithm for continuous control tasks \cite{lillicrap2015continuous} that has been widely successful in data-rich applications.
    \item \textbf{TD3.} Twin Delayed Deep Deterministic policy gradient algorithm \cite{fujimoto2018addressing} builds on DDPG and applies various modifications to improve its stability and learning performance.
\end{itemize}
For both algorithms, we use the publicly available implementations provided in \cite{fujimoto2018addressing} along with the default hyperparameters. We also provide performance measures of three baseline agents:
\begin{itemize}
    \item \textbf{Random Agent} - a dynamics-agnostic agent which selects random amounts of fertilizer and irrigation to apply at each time step.
    \item \textbf{Standard Practice Agent} - applies predetermined amounts of inputs on scheduled days during the early, mid, and late stages of the growing season. This corresponds to traditional farming methods of applying inputs during different phases of the crop's growth.
    \item \textbf{Reactive Agent} - applies high concentrations of fertilizer and irrigation whenever nitrogen levels are depleted and the soil water content is below a certain threshold.
\end{itemize}
The implementation of the Standard and Reactive agents is based on the descriptions provided in \cite{overweg2021cropgym}.

\begin{figure}[hbtp!]
\centering
\includegraphics[width=0.4\textwidth]{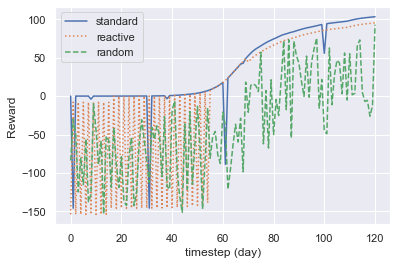} 
\caption{Performance of baseline methods on one full growing season.}
\label{fig:baseline_comparison}
\end{figure}

\subsection{Evaluation}
We measure the performance of the various agents on SWATGym. We conduct two sets of evaluations for each agent, (1) one that covers a full growing season (equivalent to one episode with 120 time steps/days) and (2) another set that is applied over multiple seasons spanning 200 episodes. The latter is repeated $5$ times, each with a different seed. In both cases, evaluations are performed every $7$ days and the results averaged over 10 runs.

\begin{figure}[hbtp!]
\centering
\includegraphics[width=0.4\textwidth]{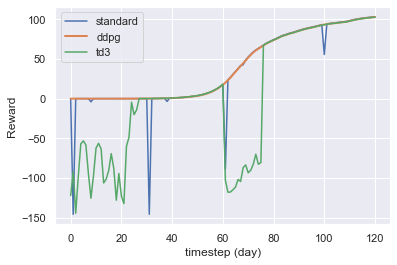} 
\caption{Performance of best baseline method against RL methods on one full growing season}.
\label{fig6}
\end{figure}

\paragraph{Single growing season.} 
We train each agent for a full corn growing season, equivalent to 120 days. Figure \ref{fig:baseline_comparison} shows the outcomes of the three baseline agents. The standard approach obtains the best performance overall compared to that of the Random agent and Reactive agent, which is an expected outcome given this strategy is widely used in practice. This comparison also highlights one way in which SWATGym can be used to evaluate management strategies. With the Standard agent as the best baseline, we also evaluate its strategy against the RL-based strategies of the DDPG and TD3 agent (Figure \ref{fig6}).

\begin{figure}[hbtp!]
\centering
\includegraphics[width=0.4\textwidth]{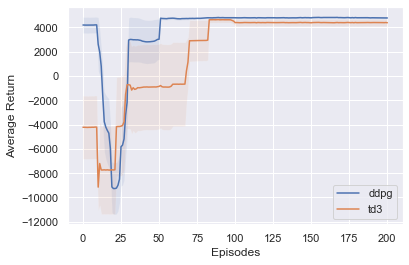} 
\caption{Learning curves of the RL algorithms on SWATGym. The shaded area represents half a standard deviation of the average evaluation over 10 runs of the experiment, each with a different seed.}
\label{fig:rl_perforamance}
\end{figure}

We observe that the DDPG algorithm performed better than both the Standard Agent and the TD3 agent on this task. This is interesting in itself because TD3 is an improvement of DDPG. We provide the learning curves of both algorithms in Figure \ref{fig:rl_perforamance}. 

\begin{figure}[ht]
\centering
\includegraphics[width=0.4\textwidth]{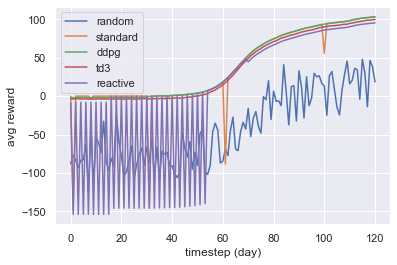} 
\caption{Average reward across 5 runs for all agents.}
\label{fig:all_agents}
\end{figure}

\paragraph{Multiple growing seasons.}
We also compute the average return ($\pm$ one standard deviation) across 5 runs of the experiment, each with a different seed. Table \ref{table:performance} shows the maximum average return performance of the agents on SWATGym and Figure \ref{fig:all_agents}) shows the average reward for all agents across 5 repetitions of the experiment. In comparison to the baseline agents, the DDPG and TD3 results illustrate the potential of RL in revolutionizing crop management strategies.

\begin{table}[!htbp]
\centering
   \small
   \begin{tabular}{lr}
   \toprule
   \textbf{Method} & \textbf{Performance}\\ 
   \midrule
   Random & $-4547.152 \pm 146.386$  \\
   Standard & $4396.402 \pm 13.116$  \\
   Reactive & $279.121 \pm 32.234$\\
   DDPG & $4169.202 \pm 2249.513$\\
   TD3 & $3344.010 \pm 4519.612$\\
   \bottomrule
   \end{tabular}
   \caption{Maximum average return $\pm$ one standard deviation for all agents across 5 repetitions of the experiment.}
\label{table:performance}
\end{table} 

\section{Discussion and Research Directions}
Beyond highlighting the potential of RL strategies in facilitating sustainable crop management, our preliminary results demonstrated the value of SWATGym as a benchmark framework. In this section, we outline key challenges in modeling crop growth and briefly describe social challenges such as the adoption of RL-based crop management strategies. 

\paragraph{Challenges in crop modeling.} Accurately modeling crop growth is a key and active research problem. SWATGym has lots of room for improvement to increase its fidelity. Yet, one must also balance simplicity and generalizability to different regions, climates, crop types, and soil characteristics. Future releases of the framework will extend the dynamics to include more processes such as groundwater seepage and other applications such as offline RL.

 \paragraph{Seasonal spatial-temporal variability.}
Most crops that are simulated are short-season crops. Regardless of the season duration, it is important to capture the spatial-temporal variability of the season. For example, evapotranspiration is higher in the summer than in winter, so the crop management algorithm should apply more irrigation to match water loss during that period. Likewise, surface runoff could be higher in the spring or rainy season (e.g., when the snow melts/after it rains). As such, the best strategy might be to apply less fertilizer to avoid leaching and waste or to apply it early in the growing season before runoff affects it.

\paragraph{Reality Gap.}
SWATGym simulates crop growth daily and allows agents to select inputs on any given day. In practice, such operations are done over a week or more, depending on the size of the field and the type of equipment used to irrigate or fertilize. Furthermore, the environment can be further constrained to limit the total amounts of inputs applied throughout the growing season and terminate whenever this threshold is reached. RL developments must be done in conjunction with domain experts to ensure the applicability of ML recommendations.

\paragraph{Social Impact.}
Through IBM's Sustainability Accelerator program, a pro bono social impact program that leverages technologies such as hybrid cloud and AI to enhance and scale non-profit and government organizations, plans are underway to deploy this framework in small-holding farms in arid regions of the United States. The goal is to help farmers make better decisions about crop management and offer a platform where they can easily evaluate different strategies to optimize crop production. This highlights a challenging aspect of this work, RL, and digital solutions to real-world problems: adoption. Working with small-holding farmers will present unique opportunities to learn how to translate research ideas into useful real-world products.

\paragraph{Climate Change.}
The negative impacts of climate change are already being felt, in the form of increasing temperatures, weather variability, shifting agroecosystem boundaries, invasive crops, and pests, and more frequent extreme weather events \cite{calzadilla2013climate}. ML-informed decision-making must be responsive to both shifting spatial and temporal patterns that influence decision-making, as also the increased frequency of extreme weather events. Approaches that improve robustness to these events and allow farmers to better understand the implications of different actions in a more volatile environmental system are critical to improving adaptation to climate change

\section{Conclusion}
Evaluating and comparing different crop management strategies in the real world is a costly and time-consuming task. Simulated environments offer a more efficient solution, allowing for the simultaneous benchmarking of multiple strategies with minimal cost. To this end, we introduced SWATGym, a reinforcement learning environment designed to make it easy to simulate crop growth and evaluate crop management strategies. SWATGym models crop growth processes from emergence to harvest and can be used to evaluate crop management strategies, which can inform decision-makers and promote sustainable agriculture practices. We hope that SWATGym will facilitate follow-up work and encourage collaboration between researchers and practitioners in both reinforcement learning and agriculture to develop better crop management strategies and promote sustainable agriculture.

\bibliography{aaai23_references}
\end{document}